# Energy Optimized Robot Arm Path Planning using Differential Evolution in Dynamic Environment

Sourya Dipta Das , Victor Bain , Pratyusha Rakshit
Department of Electronics and telecommunication Engineering
Jadavpur University, Kolkata-700032, India
dipta.math@gmail.com, victor.bain14@gmail.com, pratyushar1@gmail.com

*Abstract*— **Robots are widely used in industry due to their efficiency and high accuracy in performance. One of the most intriguing issues in manufacturing stage of production line is to minimize significantly high percentage of energy consumed by these robot manipulators. The energy optimal control of robotic manipulators is a complex problem, as it requires a deep understanding of the robot's kinematics and dynamic behaviors. This paper propose a novel method of energy efficient path planning of an industrial robot arm in a workspace with multiple obstacles using differential evolution (DE) algorithm. The path-planning problem is formulated as an optimization problem with an aim to determine the shortest and energy optimal path of the robot arm from its given initial position to the pre-defined goal location, without hitting obstacles. Application of such evolutionary algorithms in trajectory planning is advantageous because the exact solution to the path-planning problem is not always available beforehand and must be determined dynamically. Experiments undertaken reveal that the DE-based path-planning algorithm outperforms its contenders in a significant manner.**

## I. INTRODUCTION

Many industrial countries witnessed an increase in prices of both electricity and fuel during the last decade. According to the recent statistics one of the large consumers of energy is manufacturing industry. The majority of the energy is usually consumed by robots used in the manufacturing industry. In addition, the optimal usage of energy in robots plays an important role in minimizing $CO_2$ emission in the production stage of a product's lifecycle. Industrial robots are often regarded as unsustainable equipments that demand a high level of energy consumption. On the other hand, those robots provide precision; strength and sensing capabilities which can produce high quality end products. Consequently the robotic energy consumption became a major objective for many research groups and robot manufacturers. Several researchers focused on defining tools to measure and analyses the robot's energy consumption.

Now, in path planning by robot arm for object avoidance, there are many research work like genetic algorithms (GA) have been used in this field. Tian and Collins et al. [3] have proposed a method for a planar robot arm motion planning using GA. However only point obstacles have been considered and the environment is entirely two dimensional. Thus scope of practical implementation is limited. Other works in this field include [4] and [5]. Neural networks have also been used by some researches to solve the problem of obstacle avoidance in motion planning. Ziqiang Mao and T.C. Hsia et al. [5] employed neural networks to solve the inverse kinematics problem of redundant robot arms. In this paper, we propose a method to solve the problem of energy efficient path planning without obstacle collision using differential evolution (DE) algorithm proposed by Storn and Price in 1996 [2].

DE mimics the Evolution process of nature. It is a Simple and efficient adaptive scheme for global optimization over continuous spaces. This Stochastic, population-based optimization algorithm is very famous among researchers because of its reduced computational cost &simplicity. A suitable cost function has been designed according to the problem statement which has been minimized to track the optimal path between an initial and final configuration of the robot arm joints. Our algorithm has been described for a 3 DOF robot arm but can be easily extended to other arm configurations.

## II. PROPOSED WORK

### A. The Robot Arm

In the following Section, we use a robot arm with three links and three joints, each with one degree of freedom. Their movements are described by respective angles $t_1^0$, $t_2^0$ and $t_3^0$ as shown in Fig. 1. Lengths of sections of robot arm between two consecutive joints are $L_1$, $L_2$, and $L_3$.

- The first section of length L1 has an angular displacement or angular velocity through positive z-axis. Here, t1 is the angle of the corresponding joint.

- The second section of length L2 moves in the vertical plane respect to z axis which is described by t2 measured from the +ve z-axis. Here, t2 is joint angle of that joint.

- The third section of length L3 moves in the vertical plane also respect to z axis which is described by (t2+t3) measured from the +ve z-axis. Here, t3 is joint angle of that joint.

The arm ends in a gripper or end effector whose co-ordinates are given by

X= Sin (t1)*(L2*Sin (t2) +L3*Sin (t2+t3))

Y=Cos (t1)*(L2*Sin (t2) +L3*Sin (t2+t3))

Z=L2*Cos (t2) +L3*Cos (t2+t3)

So, here is the forward kinematics of the 3 Dof robot arm. Here, end effector consider as a point for the simplification and approximation.

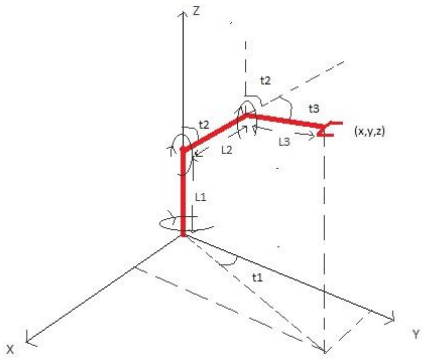

**Fig. 1.** Joint angels of robot arm

*B. Determination of intermediate Points*

Here we determine a set of intermediate points in between the initial position and the goal by using Differential Evolution Algorithm. No of intermediate points is fixed by a parameter which also depends on the constraints of the angular velocity of the motors of joints in the robot arms. Maximum angular velocity of the joint determines the minimum value of that parameter. These intermediate points makes a smooth trajectory path according to the constraints of continuous angular velocities and energy minimization.

*1) I. Kinematics of Robot Arm*

Through the intermediate points robot arm end effector reaches its goal position. Now, angular velocity, position of the joint of robot arm have to be calculated. Enumeration and equation of position of joints of robot arm is already shown above. Now, angular velocity of the joints of the arm is $\omega = d\theta/dt$ .here, it is assumed that one unit time is needed to move the end effector from initial intermediate point to next intermediate point. How much time is needed to reach the goal is depends on the actuators of the robot arm. So, parameter which is determines the number of the interior point, is the total time. So, angular velocity of the joints became numerical differential.
Finally, angular velocity is

$\omega = \frac{d\theta}{dt} \approx \Delta\theta = |\theta_f - \theta_i|$ (as $\Delta t \to 1$)
$\theta_f$ = final value of joint angle
$\theta_i$ = initial value of joint angle

As, increment of the time is one unit time, new position of non-static obstacles is updated by using the velocity vector. Those values of angular velocity, $\omega$ is used for further calculation other purposes in that intermediate point. So, value of the cost function should have variation over the distance between the current and goal position. Here, we compute the Euler distance between current intermediate point locations to goal position. So, till now cost function have the Euler distance function.so the cost function is

$$F = \sqrt{(X_g - X_i)^2 + (Y_g - Y_i)^2 + (Z_g - Z_i)^2}$$
$F_{distance} = D_i$

Where,
$X_g$ = goal point position.   $D_i$ = distance between goal point position and Current point position at i-th intermediate point.
$X_i$ = Current point position.

*2) Energy Consumption Computation*

At any point the mechanical energy of the arm will be given by the summation of kinetic and potential energies of each of the arm sections as shown below. That E is the amount of the consumed mechanical energy. So, after calculation, potential energy's value is

$$E = \sum_i (K.E._i + P.E._i)$$

In the cost function, we use relative potential difference between two consecutive interior points of the path. So, in the cost function of the optimizer, total energy is sum of kinetic energy at that interior point and difference between that intermediate point's potential energy and its previous intermediate point's potential energy.in that way, optimizer optimizes differential potential energy not absolute potential energy at that point. If total potential energy at that intermediate point is taken into cost function, then optimization may stuck at that point as to minimize the potential energy having large value, it damp down the change of the joint angle. Then, error for the deviation from the goal point will be large. That why, differential potential energy is taken to boost up the optimization process.as kinetic energy is depends on the change of joint angles, K.E. is just added to the cost function without any modification. So, mechanical energy contribution in the cost function is shown below.

$F_{energy} = |P_i - P_{i-1}| + K_i$
$K_i$ = Kinetic Energy at i -th intermediate point.
$P_i$ = Potential Energy at i-th intermediate point.

So, energy optimization is done with this part of the cost function which is shown above. Here, discussion on energy consumption and optimization for the path planning of robot arm ends.

*3) Obstacle Avoidance in Dynamic Environment*

In case of robotic arm movements, Obstacle avoidance in a dynamic environment is a lot more complex proposition than in the case of static point objects. This is because in the current situation we must not only make sure that the end effector, i.e. the gripper avoids collision but also ensure that in all its orientations all the parts of the arm avoid collisions with the obstacles.at first we build a mathematical model based on 3d geometry and vector algebra. Obstacles having various shapes can be modeled in to sphere of radius R, and center c which is circumsphere of that obstacle. So, at the beginning of the program, that radius of circumsphere, its radius, its velocity vector is initialized. As time is incremented by 1 unit after passing an interior point, position of that obstacle is updated using the velocity component of the obstacles across the x, y, z axis.

$C_j(t+1) = C_j + t* V_j$

$C_j(t)$ = Coordinate of center of circumsphere of the j th obstacle at the t-th time interval or iteration.
$V_j$ = Velocity Vector having velocity component across x, y, z axis of the j -th obstacle .

After calculation of updated coordinate of the obstacle centers, coordinates of the robot joints are also computed using forward kinematics of the robot arm. Now, vector between two consecutive joint position points is evaluated. Vector between center of circumsphere and joint point position & base point of robot arm. After that, Cross product between A and B is computed.in that way, I can find perpendicular distances from each Segments (Parts of the arm which situated between two consecutive joint point) of robot arm. Calculation is shown in the Figure -2.

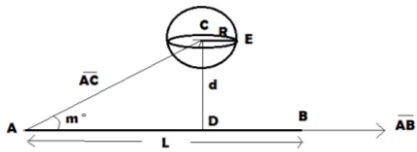

Figure: 2

$$d = \frac{|\overrightarrow{AB} \times \overrightarrow{AC}|}{|\overrightarrow{AB}|} \quad , |\overrightarrow{AB}| = L$$

Where, d= distance between obstacle and robot arm segment. This operation will be performed for each segment of the Robot arm. Now, value of the 'd' decides whether it collides with segments of the arm or not. We use a set of user defined threshold values for each segments, which decides chance of collision with arm and the obstacle. Values of the threshold is decided is depended on our precision and availability of free space to move. If value of 'd' is less than that threshold's threshold value ,then it is considered to be a threat for collision between obstacle and that arm segment. So, in the cost function it will return a constant $A_{vd}$.This constant is the sum of $n_f$ and 'd' , here,$n_f$ is a user defined constant which have a large value like 10000.this constant $n_f$ signifies that a value of d for any segment of the arm is less than the threshold so, there may be a collision. Value of $n_f$ is added with $A_{vd}$ for each value of $d_i$ the i-th segment of the arm which have less value than that segment's threshold value $Th_i$ .Value of $d_i$ is also added to the $A_{vd}$.for better precision. So, value of $A_{vd}$.is shown below.

$$A_{vd} = \begin{cases} n_f + d_i & For, d_i < Th_i \\ 0 & For, d_i => Th_i \end{cases}$$

So, value $A_{vd}$ loaded the cost function when $d_i$<$Th_i$. So, to optimize the cost function, optimizer discarded that set of joint angles which provokes such treat for the collision. In that way, Obstacle avoidance is done using the cost function of the optimizer. But, here is a Problem for which it may discard some joint angles though those joint angles may not provoke a collision between obstacle and segments of the robot arm. That case is shown in the figure-3 pictorially.

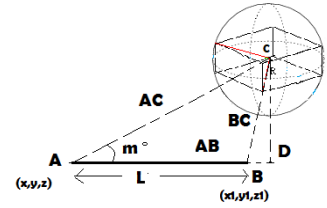

Figure-3

Here, it clear from the diagram that if $d_i < Th_i$ then according to declared model, it will be considered as a threat for the collision with the arm segment. So, for this case it is false alarm to the controller of the robot arm about collision detection. This may give bad optimization and also improper results. So, to prevent this I improved the previous model. Now, value of the perpendicular distance from the obstacle and the robot arm segment is calculated. Now, then distance from the each obstacle to the joint points of the robot arm and its base point will be evaluated. Then, value of AD can be evaluated from the Pythagoras theorem, from the figure-3 & figure-4.Now,$L_i^1$ is value of the AD and $L_i^2$ is value of BD from the figure 3 and figure 4.

$$AD = \sqrt{(AC)^2 - d_i^2}$$

AD = distance between joint point and D point.
AD = $L_i^1$, BD = $L_i^2$
$d_i$= Perpendicular distance from the center of the sphere.
AC = distance between joint point and center of the sphere.
Now, for a particular segment of the robot arm, sum of the $L_i^1$ & $L_i^2$ is evaluated to check whether it is equal to the length of that segment of the arm. if it is equal to the that length of the segment of the arm, then collision may happen with that segment with obstacle for $d_i$<$Th_i$ condition, but if sum of $L_i^1$ & $L_i^2$ is not equal to that length of the segment of the arm, then collision will not happen with only that segment with obstacle for $d_i$<$Th_i$ condition. The value of $A_{vd}$ is depend on that same condition as the previous model. In the figure-4, this modification can be understood more easily.

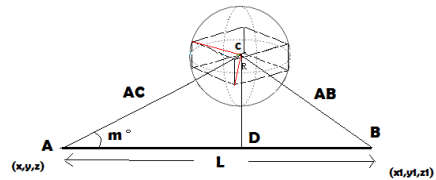

Figure – 4

So, for ith segment of length, $L_i$ of the robot arm and center of circumsphere of obstacle at C point and having same point name notations, value of $A_{vd}$ is after this modification is shown below.

$$A_{vd} = \begin{cases} n_f + d_i & For, d_i < Th_i \text{ and } L_i^1 + L_i^2 = L_i \\ d_i & Otherwise \end{cases}$$

So, cost function for the contribution of obstacle avoidance is

$F_{obstacle} = A_{vd}$.

Now, final cost function for the optimizer is following.

$F_{final} = |P_i - P_{i-1}| + K_i + D_i + A_{vd}$.

Now, we present the Differential Evolution briefly as an optimization technique of the problem.

## III. THE OPTIMIZATION TECHNIQUE

Here we have used differential evolution as our optimization technique. A cost function is designed to get the solution of the problem. Differential evolution (DE) is a powerful evolutionary algorithm developed on the framework of genetic algorithm and inspired by Nelder-Mead Simplex method . It is a simple yet efficient population-based algorithm to solve global optimization problems. It generates new individuals based on the individual difference to achieve the evolution of population. The algorithm is described below.

1. Initialization

Differential Evolution (DE) is a parallel direct search method which utilizes NP

D-dimensional parameter vectors.

$$\{\mathbf{x}_i = (x_{i,1}, x_{i,2}, ..., x_{i,D}) | i = 1, 2, ... NP\}$$

As a population for each generation G. NP does not change during the minimization process. The initial vector population is chosen randomly and should cover the entire parameter space. After initialization, DE enters a loop of Evolutionary operations: mutation, crossover and selection.

2. Mutation

For each target vector $x_i^G$ (i=1,2,...,NP, G is the current generation index), a mutant vector is generated as shown below.

$$\mathbf{v}_i^G = \mathbf{x}_{r1}^G + F \cdot (\mathbf{x}_{r2}^G - \mathbf{x}_{r3}^G)$$

Where r1, r2 and r3 are chosen randomly from the index set { 1, 2,...,NP } and different from I, and F is the scaling factor. F is a real and constant factor in the range, [0, 2] which controls the amplification of the differential variation.

3. Crossover

A binomial crossover operation operates on the target vector and mutant vector to form the trial vector $u_i^G$

$$\mathbf{u}_{i,j}^G = \begin{cases} \mathbf{v}_{i,j}^G & rand_j(0,1) \leq Cr \text{ or } j = j_{rand} \\ \mathbf{x}_{i,j}^G & \text{otherwise} \end{cases}$$

where jrand is an integer randomly chosen from 1 to D and generated for each i , and Cr is the crossover probability.

4. Selection

The selection operation selects the better one from the target vector $x_i^G$ and the trial vector $u_i^G$ according to their fitness value, and the better one will become a member of the population in the next generation.

$$\mathbf{x}_i^{G+1} = \begin{cases} \mathbf{u}_i^G & f(\mathbf{u}_i^G) \leq f(\mathbf{x}_i^G) \\ \mathbf{x}_i^G & \text{otherwise} \end{cases}$$

## IV. COMPLETE ALGORITHM FOR OPTIMIZATION PROBLEM OF PATH PLANNING

In this section solution to the discussed problem of the path planning using DE. Here angles of joints of robot arm are considered to be optimizing parameters of each solution. The complete algorithm outlining the scheme is discussed below.

**Pseudo Code:**

**INPUT:** Initial position $(x_i, y_i, z_i)$, goal position $(x_g, y_g, z_g)$ and all parameters and constraints of robot arm, other user defined parameters like $Th_i, n_f$.

**OUTPUT:** energy optimized joint angles of robot arm for the intermediate points of path planning from initial point to goal position.

**Procedure of main program**
**Begin**
Initialization of initial joint angles, robot arm's parameters;
  **For** i= 1: Time **do**
    Intermediate point $(t_f)$ = DEOptimization $(t_i, t_f, t)$;
    $t_i = t_f$;
**End for**
**Return** $t_f$;
**End**

**Procedure of DEOptimization** $(t_i, t_f, t)$
**Begin**
Initialize the all parameters and upper and lower bound of the parameters;
Generate initial population (N) of $t_f^i$;
**For** j= 1: N **do**
Angular velocity $(w_j^i)$ = difference of $t_f$ & $t_i$ ;
**End for**
**Do**
For each individual j in the population
Choose 3 number $t_1, t_2, t_3$ that, $1 <= t_1, t_2, t_3 <=N$ & $t_1 != t_2 != t_3$;
Generate random integer $I_{rand} \in (1,2,3 ..... D)$;
**For** each parameter i in the population
$v_i{}^{G+1} = tr_1{}^G + F(tr_2{}^G - tr_3{}^G)$;
**If** (randj,i ≤ CR or j = Irand)      // i = 1, 2, . . . ,N &
                                        // j = 1, 2, . . .D.
$u_{ji}^{G+1} = v_{ji}{}^{G+1}$;         // Irand is a random integer from
                                        //  [1, ...,D] & randj,I = Unirnd(0,1);
**else**
$u_{ji}^{G+1} = t_{rj,I}{}^G$ ;
**End for**
Replace current generation of $t_f$ with next generation of $t_f$f if value cost function (costfunc($t_f^g, t_i, t_g, t$))of next generation

of $t_f$ is less than the cost function value of the current generation $t_f$

**Until** termination condition is achieved;
**End**
**Return** best $t_f$;

**Procedure of costfunc**($t_f^g$, $t_i$, $t_g$, $t$)
**Begin**
Initialize Number of obstacle, its centers, radius, and velocity components;
Updates position of obstacle position with its velocity component and t (time);
Evaluate of end effector position;
$A_{vd}$= Collisioncheck(x,R,C,w);
$P_f$= Evaluate Potential energy of current position using joint angles $t_f$;
$P_i$= Evaluate Potential energy of previous position of current position using joint angles $t_f$;
D=Distance from the goal position and current position;
K=Evaluate Kinetic energy at current position using angular velocity of joints of the robot arm;
f=|$P_f$-$P_i$| +K+$A_{vd}$+d;
**Return** f;
**End**

**Procedure of Collisioncheck(x,R,C,w)**
**Begin**
Initialize C, $Th_i$, $n_f$, $d_i$;
H=Evaluate perpendicular distance from center of the obstacle to each segment of robot arm;
Z= Evaluate distance from the joints and base of the robot arm to the center of the obstacles;
Compute value of $L_i^1$= AD, $L_i^2$= BD for each obstacles;
**if** ( $d_i$<$Th_i$&& $L_i^1$+$L_i^2$=Li )
$A_{vd}$= $n_f$+$d_i$;
**else**
$A_{vd}$= $d_i$;
**Return** $A_{vd}$;
**End**

## V. EXPERIMENT AND COMPUTER SIMULATION

We implemented the algorithm in MATLAB software. Here is a test run to see if we can reach its goal by following the procedure. The Experiment was performed with a Virtual PUMA Robot Arm. For the simulation, starting and goal positions are predefined prior to initiating the experiment. Here we perform a test run to see if the arm can reach its goal by following the above procedure. The arm starts from initial position defined by ($x_i$, $y_i$, $z_i$) and reaches the goal defined y($x_g$, $y_g$, $z_g$). It should be noted here that different parameters used in the test, are not absolute measurements of any quantity but are relative values estimated to hold in proportionality. Here that for the cubical obstacle shown the edge length is s, so that R (radius) for the equivalent sphere becomes (√3/2)*s. That circumsphere of that cube is used for further calculation. From the figure-5, the plot of value of cost function is decreased over the No of iteration of DE. From the figure-6, plot of distance between end effector of robot arm and destination point is decreased and converges to zero. we have used 3 non static obstacles to perform the simulation and get average almost 98.72% accuracy to reach the goal point with error less than 0.1. In Table 1, simulation parameters are shown .Those parameters are chosen to get the simulation as much realistic as possible. In table 2, parameters for the Optimizer is shown. These parameters have been developed after many trials and have been found to give better performance.

| Size of the Population ( N ) | 150 |
|---|---|
| No of real parameters ( D ) | 6 |
| Upper Bound of the Parameter ( Smax) | 160 110 135 |
| Lower Bound of the Parameter ( Smin ) | -160 -110 -135 |
| Maximum Iteration (Maxit) | 100 |
| Probability CR ( CR ) | 0.96 |

TABLE II: PARAMETERS OF THE OPTIMISER

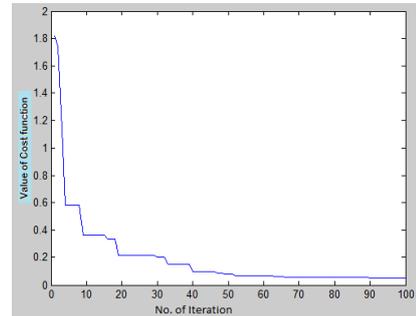

Figure-5: plot of values of the cost function over No. of iteration

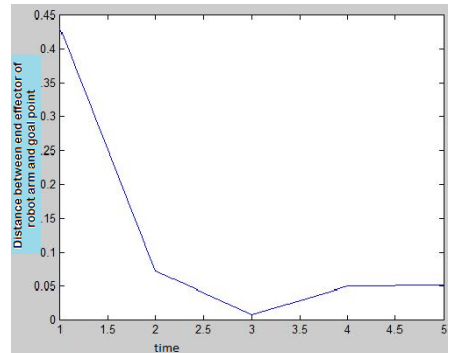

Figure-6: plot of the distance between end effector of robot arm and goal point

## VI. RESULTS AND DISCUSSION

The converging speed of DE is good.it might encounter the premature convergence in optimizing multimodal function. More fast convergence can be achieved by choosing more suitable constant parameters of the optimizer. The value of the cost function is decreased ell over the iteration, so cost function minimizes properly for each interior point. The decreasing plot of distance between end effector and goal point and less error for the deviation of final position of end effector from goal position shows this algorithm is efficient and have a high precision and accuracy.by taking suitable parameters, this accuracy can be improved further. This algorithm is also flexible to the dynamic environment. Precision and safety levels for the robot arm can be adjusted by the user according to environment and distribution of the obstacles. It is also shown that part of the cost function for obstacles avoidance in dynamic environment is not also loaded cost function when there is no chance of collision with obstacles and robot arm's segments. So, it gives more liberty to the cost function for more optimization of the energy consumption and distance between end effector and goal position. Now, it can be concluded that DE performs its role of optimization for energy optimization with obstacle avoidance perfectly.

## VII. CONCLUSION

The algorithm determines a set of optimized intermediate points for obstacle avoidance. The obtained interior points ensures smooth motion of the end-effector and optimizes energy consumption during motion from one via interior point to the next. As we have already seen above results, model simulations show that the DE algorithm can be successfully used as an optimization tool in path planning and obstacle avoidance in dynamic environment for robotic arms. It is an effective optimization tool that can be applied to give successful results in this scenario. This process helps in dynamically determining the most suitable path subject to conditions in the surroundings. Favorable results from simulations call for further research in this area. There is scope for adopting such algorithms in real robots and environments for better efficiency, and in cases of further complexities, for example on cases where there are larger number of joints and multiple static and non-static obstacles. Future research work will involve studying the mathematical effects of the optimizer parameters on the efficiency of the problem. Further analysis can be done on the applicability of other popular evolutionary algorithms in the path planning of the robotic arm for energy optimization in dynamic environment and studying their relative performance.

## REFERENCES

[1] Smetanová, A. Optimization of energy by robot movement. MM Science Journal, (March), pp. 172–173, 2010.

[2] Storn, Rainer and Kenneth V. Price. "Differential Evolution - A Simple and Efficient Heuristic for global Optimization over Continuous Spaces." J. Global Optimization 11 (1997): 341-359.

[3] Tian, Lianfang, and Curtis Collins. "An effective robot trajectory planning method using a genetic algorithm." Mechatronics 14.5 (2004): 455-470.

[4] D. Pack, G. Toussaint and R. Haupt, " Robot trajectory planning using a genetic algorithm," SPIE 1996;2824,pp.171-182.

[5] Z. Mao and T.C. Hsia, " Obstacle avoidance inverse kinematics solution of redundant robots by neural networks," Robotica (1997), vol.15, pp.3-10, 1997 Cambridge University Press.

[6] Y Saab and M. VanPutte, "Shortest path planning on topographical maps," IEEE Transactions on Systems, Man, and Cybernetics–Part A 1999;29(1) ,pp.139-150.

[7] E.G. Gilbert and D.E. Johnson , "Distance function and their application to robot path planning in the presence of obstacles," IEEE J. of Robotics and Automation RA-1(1), pp.21-30 (1985).

[8] O. Khatib, "Real-time obstacle avoidance for manipulators and mobile manipulators," Int. J. of Rob. Res., vol. 5, no. 1, pp. 90–98, 1986.

[9] Mahjoubi, H & Bahrami, Fariba & Lucas, C. (2006). Path Planning in an Environment with Static and Dynamic Obstacles Using Genetic Algorithm: A Simplified Search Space Approach. 2483 - 2489. 10.1109/CEC.2006.1688617.

[10] T. C. Lai, S. R. Xiao, H. Aoyama and C. C. Wong, "Path planning and obstacle avoidance approaches for robot arm," 2017 56th Annual Conference of the Society of Instrument and Control Engineers of Japan (SICE), Kanazawa, 2017, pp. 334-337.

[11] T. Kunz, U. Reiser, M. Stilman and A. Verl, "Real-time path planning for a robot arm in changing environments," 2010 IEEE/RSJ International Conference on Intelligent Robots and Systems, Taipei, 2010, pp. 5906-5911.

[12] Y. Chen, Y. Wang and X. Yu, "Obstacle avoidance path planning strategy for robot arm based on fuzzy logic," 2012 12th International Conference on Control Automation Robotics & Vision (ICARCV), Guangzhou, 2012, pp. 1648-1653.

[13] C. Helguera and S.Zegloul, "A local based method for manipulators path planning in heavy cluttered environments," in Proceedings of the 2000 IEEE Int. Conf. on Robotics and Automation, San Fransisco, CA, USA,2000, pp.3467-3472.